\documentclass[10pt,journal,compsoc]{IEEEtran}
\newif\ifpeerreview

\peerreviewfalse

\usepackage[nocompress]{cite}
\usepackage{url}
\usepackage{amsmath,amssymb,graphicx}

\usepackage{lipsum} % Only used to generate random text.

\usepackage[switch]{lineno}

\usepackage{times}
\usepackage{epsfig}

\usepackage{gensymb}

\usepackage{float}

\usepackage{cuted}
\usepackage{capt-of}

\usepackage{xcolor}
\usepackage{tabularx}

\DeclareMathOperator*{\argmin}{arg\,min}

\newcommand{\paperID}{25}

\title{3D Face Reconstruction using \\Color Photometric Stereo with \\Uncalibrated Near Point Lights}

\author{Zhang~Chen,
        Yu~Ji,
        Mingyuan~Zhou,
        Sing~Bing~Kang,~\IEEEmembership{Fellow,~IEEE,}
        and~Jingyi~Yu,~\IEEEmembership{Member,~IEEE}% <-this % stops a space
}

\begin{document}

\IEEEtitleabstractindextext{%
\begin{abstract}
We present a new color photometric stereo (CPS) method that recovers high quality, detailed 3D face geometry in a single shot. Our system uses three uncalibrated near point lights of different colors and a single camera. For robust self-calibration of the light sources, we use 3D morphable model (3DMM)~\cite{gerig2018morphable} and semantic segmentation of facial parts. We address the spectral ambiguity problem by incorporating albedo consensus, albedo similarity, and proxy prior into a unified framework. We avoid the need for spatial constancy of albedo; instead, we use a new measure for albedo similarity that is based on the albedo norm profile. Experiments show that our new approach produces state-of-the-art results from single image with high-fidelity geometry that includes details such as wrinkles.
\end{abstract}

}

% Make Title
\ifpeerreview
\linenumbers \linenumbersep 15pt\relax 
\author{Paper ID \paperID\IEEEcompsocitemizethanks{\IEEEcompsocthanksitem This paper is under review for ICCP 2020 and the PAMI special issue on computational photography. Do not distribute.}}
\markboth{Anonymous ICCP 2020 submission ID \paperID}%
{}
\fi
\maketitle

% The first section title should be wrapped inside a \IEEEraisesectionheading as follows.
\IEEEraisesectionheading{
  \section{Introduction}\label{sec:introduction}
}
% The very first letter of the paper is a 2 line initial drop letter
% followed by the rest of the first word in caps.
\IEEEPARstart{S}{tate}-of-the-art photometric stereo solutions for 3D face reconstruction \cite{di4d,3dmd,eisko,ma2007rapid} are capable of producing movie-quality, photo-realistic results. However, these systems tend to be bulky and expensive and generally require taking multiple shots. Even with elaborate time-multiplexing, it is difficult to capture fine facial geometry movements unless using an ultra-fast speed camera coupled with high precision synchronized light sources. The light sources and cameras also require accurate calibration to avoid distortions in the final reconstruction. 

In this paper, we present a novel lightweight one-shot solution based on uncalibrated color photometric stereo method that simply uses a camera and three uncalibrated near point light sources of different color. Our approach eliminates the need of time multiplexing, and therefore can be used to recover dynamic facial motions. Compared with distant light sources which require relatively strong power, the use of near point light sources makes the system more portable by reducing the cost and space requirement. However, for near-field lighting, one needs to know the relative positions between light sources and face geometry. Even with light positions calibrated using special calibration targets (e.g., sphere and planar light probes), one still require extra depth information of the captured object. We instead propose a self-calibration method exploiting the shape prior of human faces encoded in 3D morphable model (3DMM) \cite{gerig2018morphable} and can directly self-calibrate the relative positions between light sources and face geometry with a single image.

For objects with non-gray albedo, color photometric stereo is inherently under-determined due to spectral inconsistencies of surface reflectance: albedo is not identical under different spectra and therefore there are more unknown variables than there are constraints. 
We address the spectral ambiguity problem by proposing albedo similarity and proxy prior, and incorporating them with albedo consensus into a unified framework. As a result, our approach does not need to assume spatial constancy of albedo. We also present a new measure for albedo similarity based on the albedo norm profile. 
%We show that this new measure results in significantly higher accuracy than the measures based on image intensities. 
The proposed albedo similarity and proxy prior effectively correct distortions caused by incorrect albedo consensus in prior work. Experiments show that our new approach can produce state-of-the-art results from single image with high-fidelity geometry that includes details such as wrinkles.
%\singbing{Never use words like ``extensive", because this is highly subjective. Let the reviewer decide if indeed the experiments are extensive or exhaustive enough.}

Our technical contributions are as follow:
\begin{itemize}
	\item A self-calibration method utilizing 3DMM proxy face for color photometric stereo with near point lights.
	\item A per-pixel formulation for solving normal and albedo from color photometric stereo.
	\item A framework that incorporates albedo similarity and proxy prior with albedo consensus to produce accurate 3D reconstruction.
\end{itemize}

\section{Related Work}

Structured light~\cite{zhang2008spacetime,zhang2006high} and multi-view stereo~\cite{furukawa2009dense} have been used to reconstruct faces. While they can accurately reconstruct coarse shapes, they are less successful in recovering high frequency details such as wrinkles. On the other hand, photometric stereo~\cite{woodham1980photometric} is capable of recovering high frequency details. Techniques that combine stereo and photometric stereo exist~\cite{ma2007rapid,ghosh2011multiview,gotardo2015photogeometric}, but the combination is at the expense of a complicated hardware setup. Recently, Gotardo \textit{et al.}~\cite{gotardo2018practical} achieves high-quality dynamic face reconstruction with multi-view stereo and constant white lights through an inverse rendering framework. However, they still require careful geometric and photometric calibration as well as HDR light probe of the surrounding environment.

\subsection{Photometric Stereo (PS)}
Traditional PS~\cite{woodham1980photometric} uses 3 or more distant lights (of the same color) and sequentially creates different directional illumination by turning on only one light at a time. A sequence of images is captured, each with a different light source. The surface orientation map can then be inferred from image intensities using an over-determined linear system. Normal integration is then applied to obtain a 2.5D reconstruction. We refer readers to \cite{herbort2011introduction,ackermann2015survey} for a comprehensive review of classical PS methods. The distant light requirement has since been relaxed; much work has been done using more practical near point light sources \cite{clark1992active,higo2009hand,sakaue2011new,mecca2014near,wetzler2014close,ahmad2014improved,xie2015photometric,logothetis2017semi,liao2017indoor,queau2018led,liu2018near}. Notably, Liu et al.~\cite{liu2018near} use an LED ring with a radius of only $30mm$ centered at camera lens. Alternative self-calibrating methods \cite{alldrin2007resolving,shi2010self,wu2013calibrating,lu2013uncalibrated,papadhimitri2014closed, park2017robust} provide simpler and more flexible solutions under various assumptions \cite{8281537}. It is also possible to use uncalibrated near point light sources \cite{koppal2007novel,papadhimitri2014uncalibrated,cao2017sparse,xie2019self}, but they all require sequential capture. 

\subsection{Color Photometric Stereo (CPS)}
CPS has the key benefit of acquiring only one image and hence can be directly used to reconstruct dynamic objects. Most existing approaches use red, green, and blue lights along with a color camera~\cite{drew1994closed,kontsevich1994reconstruction,woodham1994gradient}.  Hern{\'a}ndez \textit{et al.}~\cite{hernandez2007non} apply such a technique to dynamic cloth reconstruction; they use a planar board with cloth sample fixed in the center to calibrate the coupled matrix containing reflectance, camera response, lighting spectrum, and lighting directions.
Vogiatzis and Hern{\'a}ndez~\cite{vogiatzis2012self} first construct a coarse 3D face using structure from motion and then impose the constant chromaticity constraint for shape refinement. Klaudiny~\textit{et al.}~\cite{klaudiny2010high} use a specular sphere to estimate lighting directions. To ensure constant chromaticity, they apply uniform make-up to faces. Bringier~\textit{et al.}~\cite{bringier2008photometric} explicitly calibrate the spectral response of camera and assume gray color or known uniform color.
% \cite{smith2005dynamic} presented a single-shot photometric stereo
% method by using a capturing system that works in the near infrared
% region. They used the fact that materials such as paints tend to transmit
% well in the near infrared spectra and that reflected light from the
% background material is often insensitive to wavelength changes. A
% multi-colored surface painted with such materials responds to the near
% infrared light as if it were a gray-colored surface responding to visible
% light.
% \cite{Kawabata2016OneSP} also used a multispectral camera and a reflectance
% basis set obtained from principal component analysis, and
% they added a smoothness constraint on the surfaces.

To eliminate the need of constant chromaticity, there are methods \cite{de2009capturing,kim2010photometric} that combine spectral and time-multiplexing; optical flow is then used to align adjacent frames. Jank{\'o} \textit{et al.}~\cite{janko2010colour} make use of temporal constancy of surface reflectance to eliminate the need for time-multiplexing, but an image sequence is still required as input. Gotardo \textit{et al.}~\cite{gotardo2015photogeometric} simultaneously solve for color photometric stereo, optical flow, and stereo matching within each 3-frame time window, but require 9 color lights. Rahman~\textit{et al.}~\cite{rahman2014color} arrange complementary color lights on a ring. Their approach requires using 2 images under complementary illuminations as input. Anderson \textit{et al.}~\cite{anderson2011color} assume piecewise constant chromoticity by segmenting a scene into different chromaticities. To calibrate chromaticities, they also require a stereo camera pair to obtain coarse geometry.

Fyffe \textit{et al.}~\cite{fyffe2011single} extend the usual 3 color channels to 6 by using 2 RGB cameras and a pair of Dolby dichroic filters. An extension of their work~\cite{fyffe2015single} employ polarized color gradient illumination but require a complex setup with 2040 LED light sources. Chakrabarti and Sunkavalli~\cite{chakrabarti2016single} observe that the reflectance and normal within a uniform color region can be uniquely recovered from spectrally demultiplexed image by assuming piecewise constant albedo. Ozawa \textit{et al.}~\cite{ozawa2018single} densely discretize albedo chromaticity and enforce consensus on albedo norms to reconstruct objects with spatially-varying albedo. However, most of these approaches assume directional lighting and require pre-calibrating them.
It is possible to use near light sources~\cite{collins20123d}, but they still require pre-calibration. In contrast, our technique focuses on face reconstruction and exploits prior face information to enable self-calibration of near point lights. We assume unknown light positions and spatially-varying albedo. The former enables more feasible capture while the latter fulfils the physical property of real faces. 

\subsection{Single Image Techniques}
There are methods for inferring face geometry from a single unconstrained image; see \cite{zollhofer2018state} for an overview of state-of-the-art methods. However, they tend to produce less accurate results compared with multi-view stereo and photometric stereo. Piotraschke and Blanz~\cite{piotraschke2016automated} demonstrate the usefulness of semantic segmentation to improve reconstruction quality. In our work, we use the 3D morphable model~\cite{gerig2018morphable} to obtain an initial proxy face for light source calibration. 

Shape-from-shading and deep learning based approaches have also been adopted to recover details~\cite{cao2015real,richardson20163d,sela2017unrestricted,richardson2017learning,tran2018extreme,guo2018cnn,li2018feature, Chen_2019_ICCV}. Jiang \textit{et al.}~\cite{jiang20183d} combined local corrective deformation fields with photometric consistency constraints. Yamaguchi \textit{et al.}~\cite{yamaguchi2018high} use a large corpus of high-fidelity face captures from the USC Light Stage~\cite{ghosh2011multiview} to learn the mapping from texture to highly-detailed displacement map. These solutions can provide visually pleasing results but accuracy is heavily dependent on illumination. 
% However, such data cannot be easily captured in large magnitude. 
% Our work uses a much more accessible hardware setup, which can easily be used to collect large amount of 3D face data.
%\singbing{Can your method reconstruct facial hair or unusual facial features?}
%\zhang{Yes, we will show some close up of eyebrows in real experiment.}
%\singbing{Details are already ``fine," so ``fine details" is redundant.}

\section{Color Photometric Stereo with Near Point Lights}

Traditional color photometric stereo uses 3 distant lights with different lighting directions and spectrum (usually red, green and blue) together with an RGB camera to spectrally multiplex different illumination in a single image. By assuming distant lights, each surface point is illuminated by three directional lighting with direction $\mathbf{l}_{j}\in\mathbb{R}^3$ and spectral distribution $\mathcal{E}_{j}(\lambda)$, where $j=1,2,3$ and $\lambda$ is the wavelength. We denote the normal and reflectance function at any pixel $(x,y)$ as $\mathbf{n}(x,y)$ and $\mathcal{R}(x,y,\lambda)$, respectively. Let $\mathcal{S}_{i}(\lambda)$ with $i=1,2,3$ be the spectral response of each camera color channel. For a Lambertian surface, the image pixel intensity $c_{i}(x,y)$ can be expressed as
\begin{equation}\label{eq:distant_color_ps_vector}
c_{i}(x,y) = \sum_{j} \mathbf{l}_{j}^\top \mathbf{n}(x,y) \int \mathcal{S}_{i}(\lambda) \mathcal{R}(x,y,\lambda) \mathcal{E}_{j}(\lambda) d\lambda.
\end{equation}

We denote $\mathbf{A}(x,y)\in\mathbb{R}^{3\times3}$ as the albedo matrix whose element at $i$th row and $j$th column is
\begin{equation}\label{eq:albedo_matrix}
\mathbf{A}_{i,j}(x,y)=\int \mathcal{S}_{i}(\lambda) \mathcal{R}(x,y,\lambda) \mathcal{E}_{j}(\lambda) d\lambda.
\end{equation}

Each element of $\mathbf{A}(x,y)$ thus represents the albedo under one light-channel pair. Letting $\mathbf{c}=[c_{1},c_{2},c_{3}]^\top$ and $\mathbf{L}=[\mathbf{l}_{1},\mathbf{l}_{2},\mathbf{l}_{3}]^\top$, we can rewrite Eq.~\ref{eq:distant_color_ps_vector} in matrix form as
\begin{equation}\label{eq:near_color_ps_matrix}
\mathbf{c}(x,y) = \mathbf{A}(x,y)\mathbf{L}(x,y)\mathbf{n}(x,y).
\end{equation}

Note that for distant lights, $\mathbf{L}$ is identical for all pixels. As a result, with initial coarse normal $\mathbf{n'}$, one can self-calibrate the product of $\mathbf{A}$ and $\mathbf{L}$ by assuming constant albedo or constant chromoticity \cite{vogiatzis2012self}. However, for near point lights, lighting direction is spatially-varying. By further taking into account the inverse square illumination attenuation due to distance, we obtain
\begin{equation}\label{eq:l_near_dist}
\mathbf{l}_{j}(x,y) = \frac{\mathbf{p}_{j}-\mathbf{v}(x,y)}{\left\lVert\mathbf{p}_{j}-\mathbf{v}(x,y)\right\rVert_2^3} ,
\end{equation}
where $\mathbf{p}_{j}$ is the 3D position of $j$th light source and $\mathbf{v}(x,y)$ is the corresponding 3D position for pixel at $(x,y)$.

\section{Near Point Light Self-Calibration}

The benefits of self-calibration are two-folded: first, it eliminates the need for a special calibration target (e.g., sphere and planar light probes) and the laborious procedure usually involved when calibrating near point lights; second, it can handle unexpected movements of hardware devices (e.g., light sources), making the capture process more robust.
To the best of our knowledge, our work is the first to address self-calibration of near point lights under color photometric stereo.
For traditional photometric stereo with near point lights of same color, numerous self-calibration methods exist \cite{koppal2007novel,papadhimitri2014uncalibrated,cao2017sparse,xie2019self}, but these methods are not directly applicable due to more unknowns in color photometric stereo. Most relevent to our work, Cao \textit{et al.}~\cite{cao2017sparse} also exploit 3DMM for self-calibration. However, a significant difference with our work is that they resolve ill-posedness by jointly solving for all lights and require the albedo of a pixel to be identical under each light. This assumption no longer holds for color photometric stereo due to spectral inconsistencies of surface reflectance, as shown in Eq. \ref{eq:albedo_matrix}. In contrast to \cite{cao2017sparse}, we propose a RANSAC-based approach in this paper.

In order to self-calibrate the light source positions, we first require a coarse proxy mesh, from which we obtain initial rough estimates for normal $\mathbf{n}$ and position $\mathbf{v}$ at every pixel $(x,y)$. Unlike other methods that use multi-view stereo \cite{vogiatzis2012self} or stereo matching \cite{anderson2011color} to obtain the proxy mesh, our approach makes use of the 3D morphable model (3DMM)~\cite{gerig2018morphable} and needs only one image as input. To compensate for the inaccuracies in the proxy mesh, we use RANSAC followed by hypothesis merging to robustly estimate light source positions. We provide details of our method in the following two sections.

\subsection{Proxy Mesh Generation}\label{subsec:proxy_mesh}
3DMM is a deformable template for the mesh of a human face. It consists of Principal Component Analysis (PCA) linear basis along three dimensions: shape, expression, and albedo. Since we are concerned with only shape and expression associated with the proxy mesh, we omit the albedo dimension. 3DMM interprets the face mesh $\mathbf{m}\in\mathbb{R}^{3n}$ as a linear combination of shape and expression bases:
\begin{equation}\label{eq:3dmm_linear_comb}
\mathbf{m} = \mathbf{a}_s + \mathbf{a}_e + \sum_{i}\alpha_{i}\mathbf{b}_i^s + \sum_{i}\beta_{i}\mathbf{b}_i^e,
\end{equation}
where $\mathbf{a}_s,\mathbf{a}_e\in\mathbb{R}^{3n}$ are PCA means and $\mathbf{b}_i^s,\mathbf{b}_i^e\in\mathbb{R}^{3n}$ are $i$th PCA bases of shape and expression, respectively. $n$ is the number of mesh vertices, and $\alpha_{i},\beta_{i}$ are $i$th coefficients for linear combination of the bases. We adopt the \emph{Basel Face Model 2017}~\cite{gerig2018morphable} for 3DMM, and use the iterative linear method from \cite{Huber2016AM3} to jointly solve for PCA coefficients and camera parameters (intrinsics and extrinsics). We then rasterize the generated proxy mesh to recover initial normal and 3D position for each pixel. While the proxy mesh resembles a human face with a reasonable pose, its geometry is usually inaccurate.

\begin{figure}[t]
\begin{center}
   \includegraphics[width=0.985\linewidth]{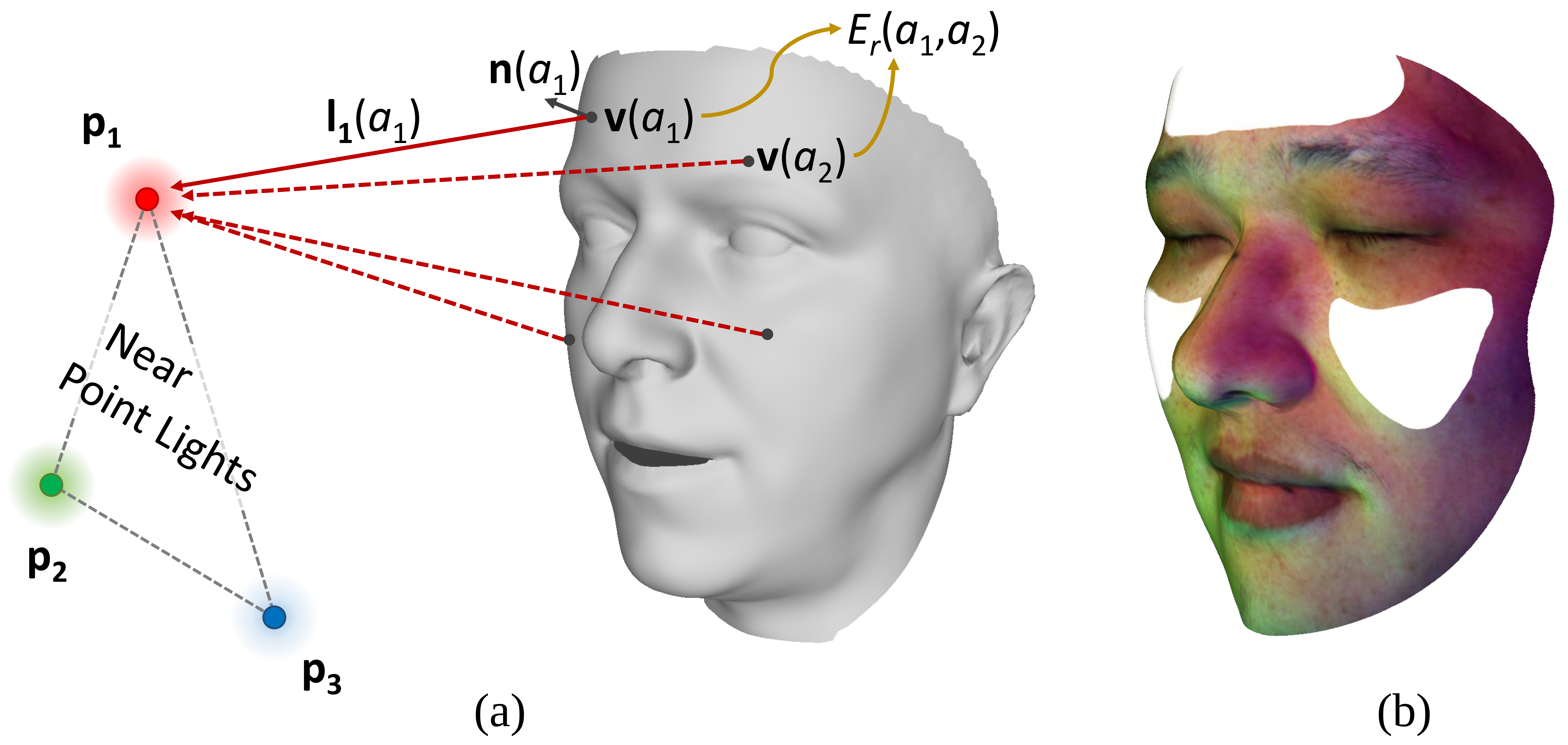}
\end{center}
   \caption{Self-calibration of near point light positions using a proxy face. (a) Parameters involved in estimating $\mathbf{p}_{1}$. (b) Regions (white) on the face used for RANSAC pixel sampling.}
\label{fig:method_self_calib}
\end{figure}

\subsection{Estimation of Light Source Positions}\label{subsec:estimate_light_pos}
As with \cite{chakrabarti2016single,ozawa2018single}, we assume that there is no crosstalk between light sources and camera channels, i.e., the spectrum of each light source can only be observed in its corresponding camera channel. As a result, the albedo matrix $\mathbf{A}(x,y)$ is diagonal. For simplicity, let $\boldsymbol{\rho}(x,y)=[\mathbf{A}_{1,1}(x,y), \mathbf{A}_{2,2}(x,y), \mathbf{A}_{3,3}(x,y)]^\top$, Eq.~\ref{eq:near_color_ps_matrix} then becomes
\begin{equation}\label{eq:near_color_ps_matrix_simple}
\mathbf{c}(x,y) = \boldsymbol{\rho}(x,y)\odot\mathbf{L}(x,y)\mathbf{n}(x,y),
\end{equation}
where $\odot$ is the Hadamard product operator. For two pixels $(x_1, y_1),(x_2, y_2)$ with equal albedo in the $i$th channel, i.e., $\boldsymbol{\rho}_i(x_1,y_1)=\boldsymbol{\rho}_i(x_2,y_2)$, we have
\begin{equation}\label{eq:equality1}
\frac{\mathbf{c}_i(x_1,y_1)}{\mathbf{L}_i(x_1,y_1)\mathbf{n}(x_1,y_1)} = \frac{\mathbf{c}_i(x_2,y_2)}{\mathbf{L}_i(x_2,y_2)\mathbf{n}(x_2,y_2)},
\end{equation}
where $\mathbf{L}_i$ is the $i$th row of $\mathbf{L}$, representing the lighting direction of $i$th light source. Substituting Eq.~\ref{eq:l_near_dist} into Eq.~\ref{eq:equality1} and moving all variables to the left hand side, we obtain
\begin{align}\label{eq:equality2}
\begin{split}
&\frac{\mathbf{c}_i(x_1,y_1)\left\lVert\mathbf{p}_{i}-\mathbf{v}(x_1,y_1)\right\rVert_2^3}{(\mathbf{p}_i - \mathbf{v}(x_1,y_1))\mathbf{n}(x_1,y_1)}\\
- &\frac{\mathbf{c}_i(x_2,y_2)\left\lVert\mathbf{p}_{i}-\mathbf{v}(x_2,y_2)\right\rVert_2^3}{(\mathbf{p}_i - \mathbf{v}(x_2,y_2))\mathbf{n}(x_2,y_2)} = 0.
\end{split}
\end{align}

Once $\mathbf{n}$ and $\mathbf{v}$ are extracted from proxy mesh, we can now recover $\mathbf{p}_i$ (in the same coordinate system as proxy mesh), which has 3 unknowns. We require at least 3 constraints, which means a minimum of 4 pixels with equal albedo in the $i$th channel. Since there is no correlation between different lights or channels in Eq.~\ref{eq:equality2}, we can estimate the position of each light independently. However, since the albedo is unknown, we cannot deterministically locate pixels with equal albedo. Our solution is to employ RANSAC to randomly sample quadruplets of pixels. Since we only require each sampled quadruplet to have equal albedo in one channel, there is still a high probability that at least one sampling provides a qualified quadruplet. 

%\singbing{$E$ (spectral distribution and unbiased measure) and $S$ (spectral response and pixel similarity) are overloaded symbols. You should use different symbols to avoid confusion.}

Notice that in Eq.~\ref{eq:equality2}, the numerators have a higher order of distance between light source and surface point than those in the denominators. This biases the solution towards closer light positions. We instead use an unbiased form of Eq.~\ref{eq:equality2} to measure the residual between two pixels $a_1,a_2$:
\begin{equation}\label{eq:equality3}
\begin{split}
E_r(a_1,a_2) = &\frac{\mathbf{c}_i(a_1)(\mathbf{p}_{i}-\mathbf{v}(a_2))\mathbf{n}(a_2)\left\lVert\mathbf{p}_{i}-\mathbf{v}(a_1)\right\rVert_2}{\left\lVert\mathbf{p}_{i}-\mathbf{v}(a_2)\right\rVert_2^2}\\
&- \frac{\mathbf{c}_i(a_2)(\mathbf{p}_{i}-\mathbf{v}(a_1))\mathbf{n}(a_1)\left\lVert\mathbf{p}_{i}-\mathbf{v}(a_2)\right\rVert_2}{\left\lVert\mathbf{p}_{i}-\mathbf{v}(a_1)\right\rVert_2^2}.
\end{split}
\end{equation}

For each quadruplet $\mathcal{Q}$ (an example is shown in Fig. \ref{fig:method_self_calib}(a)), a hypothesis of the light position is computed by solving
\begin{equation}\label{eq:calib_optim}
\min_{\mathbf{p}_i} \sum_{a_k\in\mathcal{Q}}\sum_{a_l\in\mathcal{Q}-a_k}(E_r(a_k,a_l))^2,
\end{equation}
which is a squared sum of residuals between each pair of pixels in a quadruplet. We use the Levenberg-Marquardt algorithm to solve the nonlinear optimization. 

In voting for a hypothesis, a pixel $a_w$ is considered an inlier if the squared sum of residuals between it and the pixels in $\mathcal{Q}$ satisfies
\begin{equation}\label{voting_criterion}
\sum_{a_k\in\mathcal{Q}}(E_r(a_k,a_w))^2 < \tau^2,
\end{equation}
where $\tau$ is a threshold and set as $0.01$ in our experiments.

Instead of using all pixels for sampling and voting, we only use the pixels on left cheek, right cheek, and forehead, as shown in Fig. \ref{fig:method_self_calib}(b). This is to avoid potential highly non-Lambertian regions such as facial hair and shadows. The segmentation of these regions only need to done once on a 3DMM mean face, which can then be projected to different face images~\cite{cao2017sparse}.

Unlike standard RANSAC which chooses the hypothesis with the most number of inliers as the final estimate, we perform an additional filtering and merging process on all the hypotheses. The reason is that the 3DMM-based proxy mesh is inaccurate even as low-frequency geometry. As a result, the initial normals deviate from true normals at most pixels, making consensus less concentrated and potentially drifting away from the correct hypothesis. 
% If we just pick the hypothesis with highest number of inliers, we are likely to obtain an estimate with significant errors. 
Instead, we take a set of hypotheses into account to produce a more robust estimate. 

In the filtering step, we determine a plausible region for hypotheses and ignore all hypotheses outside this region. We first use the four-point algorithm in \cite{vogiatzis2012self} to produce the calibration matrix, which is the product of dominant albedo and directional lighting directions. 
% For our work, since the spectrum of each light is only observed in its corresponding camera channel, 
We then factor out the dominant albedo and extract lighting direction $\mathbf{l}_i'$ for each light by normalizing each row of the calibration matrix. Hypothesis $\mathbf{p}_i'$ (for the $i$th light source position) is dropped if it does not satisfy
\begin{equation}\label{eq:hypothesis_filtering}
\arccos{\frac{(\mathbf{p}_{i}'-\mathbf{v}_c)^\top \mathbf{l}_{i}'}{\left\lVert\mathbf{p}_{i}'-\mathbf{v}_c\right\rVert_2}} < \eta,
\end{equation}
where $\mathbf{v}_c$ is the mean 3D position of all pixels.

Eq.~\ref{eq:hypothesis_filtering} forms a cone region with half-angle $\eta$ around $\mathbf{l}_i'$; all hypotheses outside this region are ignored. We use $\eta=15\degree$ in our experiments. Subsequently, we merge the remaining hypotheses $\mathcal{P}_i$ with weighted linear combination to obtain final estimate for a light source position:
\begin{equation}\label{eq:hypothesis_merging}
\mathbf{p}_i = \frac{\sum_{\mathbf{p}_i' \in \mathcal{P}_i}w(\mathbf{p}_i')\mathbf{p}_i'}{\sum_{\mathbf{p}_i' \in \mathcal{P}_i}w(\mathbf{p}_i')},
\end{equation}
where $w(\mathbf{p}_i')$ is the number of inliers for hypothesis $\mathbf{p}_i'$. 

\section{Face Reconstruction}\label{sec::FaceReconstruction}

\begin{figure}[t]
\begin{center}
   \includegraphics[width=0.8\linewidth]{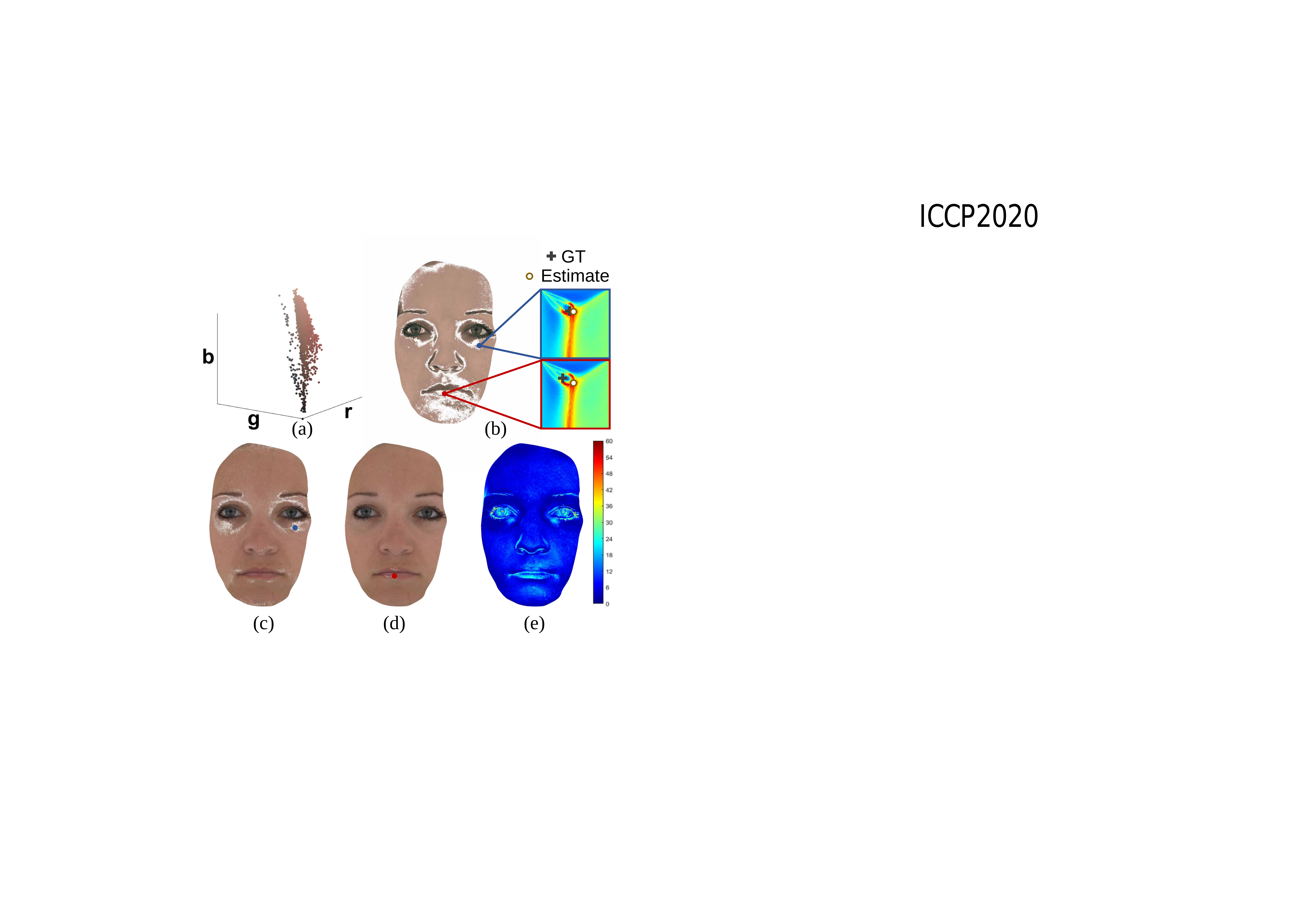}
\end{center}
   \caption{Effect of consensus term, illustrated on a face with ground truth. (a) Albedo distribution. (b) Two pixels that contribute to a consensus. The close-ups show the magnitude of negative consensus term at the two pixels in chromaticity space. The skin pixel is accurately estimated while the lip pixel is not. (c, d) Distribution of ground truth pixels that form consensus with a given pixel (indicated as blue and red dots, respectively). (e) Normal error map when using only the consensus term.}
\label{fig:method_consensus}
\end{figure}

Once the light source positions have been determined, we can obtain per-pixel lighting directions $\mathbf{L}(x,y)$ using light positions and proxy face.
We then set out to estimate per-pixel photometric normal. With the albedo unknown, this problem is pixel-wise underdetermined (from Eq.~\ref{eq:near_color_ps_matrix_simple}). This is because there are $5$ degrees of freedom ($3$ for albedo and $2$ for normal) but only $3$ constraints. It has been shown~\cite{chakrabarti2016single, ozawa2018single} that 3 pixels with equal albedo and linearly independent normals can uniquely determine the albedo and normals at these pixels. To exploit this property, Chakrabarti and Sunkavalli~\cite{chakrabarti2016single} modeled the albedo as being piece-wise constant and use a polynomial model for surface depth. However, their method tends to produce overly-smoothed results. On the other hand, Ozawa \emph{et al.}~\cite{ozawa2018single} developed an iterative voting scheme based on consensus of albedo norms to simultaneously classify pixels into different albedos and compute their normals. Since their method assumes no spatial constancy on albedo, high-frequency details can be recovered.
In the extreme case where all pixels share the same albedo, the correct albedo chromaticity can be estimated by finding the one that produces the strongest consensus on the albedo norm. However, for a multi-colored surface, their method may produce albedo consensus that leads to incorrect estimation for some pixels. 
This is because a pixel can be interpreted by any albedo chromaticity and corresponding albedo norm. There may exist situations where, under consensus albedo chromaticity, a pixel with a different albedo has a similar albedo norm with consensus. For human faces, the albedo distribution tends to spread out instead of being of a single albedo, as shown in Fig.~\ref{fig:method_consensus}a. Consensus usually arrives at a reasonable estimation for major clusters because the number of inliers tends to be large, which improves robustness. The skin pixel at the blue dot in Fig. \ref{fig:method_consensus}(b, c) shows an example. On the other hand, for minor clusters, consensus tends to provide unreliable estimation as shown by the red dot in Fig.~\ref{fig:method_consensus}(b, d), where the lip pixel can be better interpreted by an incorrect albedo chromaticity. 

By comparison, we propose a pixel-wise formulation which incorporates albedo consensus, albedo similarity between pixels as well as proxy mesh for high-quality reconstruction. 
From Eq.~\ref{eq:near_color_ps_matrix_simple}, we can decompose albedo $\boldsymbol{\rho}$ into albedo chromaticity $\hat{\boldsymbol{\rho}}$ and albedo norm $\tilde{\rho}$:
\begin{align}\label{eq:near_color_ps_matrix_transform}
\begin{split}
\mathbf{c}(x,y) = \hat{\boldsymbol{\rho}}(x,y)\odot\mathbf{L}(x,y)(\tilde{\rho}(x,y)\mathbf{n}(x,y)),\\
\tilde{\rho}(x,y)\mathbf{n}(x,y) = \mathbf{L}(x,y)^{-1}(\mathbf{c}(x,y) \oslash \hat{\boldsymbol{\rho}}(x,y)),
\end{split}
\end{align}
where $\oslash$ is the Hadamard division operator. We only need to solve for albedo chromaticity because albedo norm and normal can then be trivially computed. 

To make the problem more tractable, as with \cite{chakrabarti2016single,ozawa2018single}, we discretize albedo chromaticity in the space of positive unit sphere $\boldsymbol{S}^2_+$ into candidates $\mathcal{C} = \{\hat{\boldsymbol{\rho}}^{(1)}, \hat{\boldsymbol{\rho}}^{(2)}, ...\}$.
Then, for each pixel $a_i$, we solve for its albedo chromaticity using
\begin{align}\label{eq:face_recon:formulation}
\begin{split}
\hat{\boldsymbol{\rho}}(a_i) = \argmin_{\hat{\boldsymbol{\rho}} \in \mathcal{C}} &E_c(a_i, \hat{\boldsymbol{\rho}}) + \lambda_{s} w_s(a_i) E_s(a_i, \hat{\boldsymbol{\rho}})\\ 
&+ \lambda_{p} w_p(a_i) E_p(a_i, \hat{\boldsymbol{\rho}}),
\end{split}
\end{align}
where $E_c$ is the albedo consensus term, $E_s$ the albedo similarity term, and $E_p$ the proxy prior term. $w_s(a_i)$ and $w_p(a_i)$ modulate the influence of similarity term and proxy term at different pixels. After solving for albedo chromaticity at each pixel, we can then compute the normal and use Poisson integration to obtain geometry. Compared with proxy mesh, our final reconstruction is more accurate for both macro- (shape, expression) and micro- (wrinkles, etc.) geometries. We detail each term in the following three sections.

\subsection{Albedo Consensus}
Albedo consensus measures the number of pixels that have similar albedo norm under an albedo chromaticity candidate~\cite{ozawa2018single}. %Pixels with the same albedo must have the same albedo norm under the correct albedo chromaticity. 
To compute the consensus term, for each albedo chromaticity candidate $\hat{\boldsymbol{\rho}}^{(j)}$, we find the corresponding albedo norms of all pixels $\mathcal{N}^{(j)} = \{\tilde{\rho}^{(j)}(a_1), \tilde{\rho}^{(j)}(a_2), ...\}$ and build a histogram with bin width $\delta_b\cdot median(\mathcal{N}^{(j)})$ \cite{ozawa2018single}. Let $\mathcal{B}^{(j,k)}$ be the $k$th bin under $\hat{\boldsymbol{\rho}}^{(j)}$, $\vert \mathcal{B}^{(j,k)} \vert$ its cardinality, and $b_{i,j}$ the index for the bin that contains the albedo norm of pixel $a_i$ under $\hat{\boldsymbol{\rho}}^{(j)}$. We define
\begin{equation}\label{eq:face_recon:consensus_term}
E_c(a_i, \hat{\boldsymbol{\rho}}^{(j)}) = \frac{m - \vert \mathcal{B}^{(j,b_{i,j})} \vert}{m},
\end{equation}
where $m$ is the total number of pixels. However, it should be noted that pixels of different albedo may also have similar albedo norm under incorrect albedo chromaticities. We propose using \emph{albedo similarity} and \emph{proxy prior} to handle this problem.

\subsection{Albedo Similarity}
% \begin{figure}[t]
% \begin{center}
%   \includegraphics[width=1\linewidth]{method_similarity.pdf}
% \end{center}
%   \caption{Correlation between albedo difference (x-axis) and our similarity measure (y-axis) at four different locations. The correlation degrades at small similarity, so we downweight the similarity term at such situations. \singbing{The claim seems dubious; in A and C, the spread of albedo seems similar at different similarity measures. Am I missing something? Are the labels correct? A and C look similar and should come from similar locations, with similar arguments for B and D.}}
% \label{fig:method_similarity}
% \end{figure}

Directly inferring albedo similarity from image intensity is error-prone, since the difference in image intensity can be caused by either albedo or shading or both. Instead, the albedo norms of a pixel under all albedo chromaticities form an albedo norm profile. We reason that if two pixels have similar albedo norm profile, then they are likely to have similar albedos. From Eq.~\ref{eq:near_color_ps_matrix_transform}, letting $\mathbf{H} = [\mathbf{c}_1\mathbf{L}_{:1}^{-1}, \mathbf{c}_2\mathbf{L}_{:2}^{-1}, \mathbf{c}_3\mathbf{L}_{:3}^{-1}]$ (where $\mathbf{L}_{:i}^{-1}$ is the $i$th column of $\mathbf{L}^{-1}$) and $\hat{\boldsymbol{\rho}}' = [\frac{1}{\hat{\boldsymbol{\rho}}_1},\frac{1}{\hat{\boldsymbol{\rho}}_2},\frac{1}{\hat{\boldsymbol{\rho}}_3}] ^ T$, we have
\begin{equation}\label{eq:near_color_ps_matrix_new}
\tilde{\rho}(x,y)\mathbf{n}(x,y) = \mathbf{H}(x,y)\hat{\boldsymbol{\rho}}'(x,y).
\end{equation}

The albedo norm profile of a pixel is controlled by $\mathbf{H}$. Hence, we measure the similarity between two pixels as
\begin{equation}\label{eq:face_recon:similarity_measure}
M(a_1,a_2) = -\left\lVert \mathbf{H}(a_1) - \mathbf{H}(a_2) \right\rVert_F,
\end{equation}
where $\left\lVert . \right\rVert_F$ is the Frobenius norm.
The albedo similarity term is then computed as
\begin{equation}\label{eq:face_recon:similarity_term}
E_s(a_i, \hat{\boldsymbol{\rho}}^{(j)}) = \frac{1}{\vert \mathcal{B}^{(j,b_{i,j})} \vert}\sum_{a \in \mathcal{B}^{(j,b_{i,j})}}-M(a_i,a),
\end{equation}
which is the mean similarity between a pixel and its same-bin pixels under the $j$th albedo chromaticity candidate. 

% The correlation between our proposed similarity measure and ground-truth albedo difference on several example pixels is shown in Fig. \ref{fig:method_similarity}. 
% It shows that large similarity is a good indicator for pixels with similar albedos while the correlation degrades for small similarity. 
We further multiply a per-pixel weight to the similarity term to suppress its effect at pixels where the similarity term is large for all albedo chromaticity candidates. More specifically, we compute the weight as
\begin{equation}\label{eq:face_recon:similarity_weight}
w_s(a_i) = e^{-(min(E_s(a_i, :)) - min(E_s(:, :)))^2 / \sigma_s^2}.
\end{equation}

\subsection{Proxy Prior}
The proxy albedo chromaticity map can be computed from the proxy mesh using Eq.~\ref{eq:near_color_ps_matrix_simple} and is used to penalize implausible estimations produced by the consensus term. The proxy term is expressed as
\begin{equation}\label{eq:face_recon:proxy_term}
E_p(a_i, \hat{\boldsymbol{\rho}}^{j}) = 1 - \hat{\boldsymbol{\rho}}_p(a_i) ^ T \hat{\boldsymbol{\rho}}^{(j)},
\end{equation}
where $\hat{\boldsymbol{\rho}}_p(a_i)$ is the proxy albedo chromaticity at pixel $a_i$. We apply this term only to pixels where the consensus term gives estimations largely deviated from proxy albedo chromaticity. Otherwise, it will bias reconstruction towards the proxy mesh. We multiply the proxy term with the following per-pixel weight:
\begin{equation}\label{eq:face_recon:proxy_weight}
w_p(a_i) = e^{-(min(E_p(a_i, :)) / E_p(a_i, \hat{\boldsymbol{\rho}}_c(a_i)))^2 / \sigma_p^2},
\end{equation}
where $\hat{\boldsymbol{\rho}}_c(a_i)$ is the estimated albedo chromaticity at pixel $a_i$ using the consensus term alone.

\section{Experimental Results}

%\singbing{Make sure you show initialized shape to give the reviewers an idea of how close or far it is to the optimized shape. Does it always converge to the correct shape?}

%\singbing{You should show examples of reconstruction of 3D facial animation. Show representative frames in the paper and videos as part of supplementary material.}

%\zhang{Our system can now only capture several frames per second because the light sources are still not strong enough. I think we are not able to upgrade our system before the deadline.}

%\singbing{This is ok. You are making the claim that the technique can handle dynamic objects. Just move face/change expression a bit more slowly than usual. You need to show that movements do not cause problems with reconstruction as they would for techniques that require sequential capture and a stationary face.}

In this section, we first report results on synthetic face images generated using a high-quality face dataset and synthetic lighting. We then show results for real data captured using our setup. To self-calibrate each light, we use 2,000 iterations for RANSAC. The reconstruction parameters are set as follows: $\delta_b=0.025, \lambda_s=1.5, \lambda_p=0.5, \sigma_s=0.003, \sigma_p=0.01$. We discretize albedo chromaticity in spherical coordinates as $\{0\degree, 1\degree, \ldots, 90\degree\} \times \{0\degree, 1\degree, \ldots, 90\degree\}$. 

We compare our performance against those of representative state-of-the-art techniques \cite{vogiatzis2012self,chakrabarti2016single,ozawa2018single}. VH12 \cite{vogiatzis2012self} assumes directional lighting with single albedo chromaticity, and uses the same proxy face as our method for self-calibration. Since CK16 \cite{chakrabarti2016single} requires directional lighting directions as input, we compute approximated lighting directions as the rays from face center to ground truth light positions. OS18 \cite{ozawa2018single} originally assumes directional lighting, but we adapted it to work for near point lighting by simply using per-pixel lighting directions during computation. The per-pixel lighting directions are obtained using our estimated light positions and proxy face, which are the same as with our method. Notice that only VH12 \cite{vogiatzis2012self} and our method work under uncalibrated light sources while the other two methods require additional calibration information. After obtaining normal map, we use Poisson integration \cite{queau2015edge} to get geometry for both our method and comparison methods.

\subsection{Experiments Using Synthetic Data}
To evaluate our method objectively, we apply it to synthetic input images with known ground truth. The synthetic images are generated by rendering high-quality face data from the USC Light Stage~\cite{ma2007rapid, stratou2011effect} under near point lighting and orthographic projection, with resolution of $2048 \times 1536$. (Note that while real cameras are not based on orthographic projection, we use it in our simulations to exclude the influence of perspective and focus.)
The synthetic lights are distributed with equal azimuth angles between neighboring lights, and at the same elevation angle. The distance between each light and the face center is identical. During rendering, we retain self-shadows on the face while ignoring other shadowing effects on the background. We also avoid saturation by scaling each image so that the maximum pixel intensity is 255.

We first report our system's performance under different light distances, elevation angles, anisotropy, and crosstalk using a single face data from~\cite{ma2007rapid}. Then, we use the face dataset ICT-3DRFE~\cite{stratou2011effect} to evaluate our method for different gender, skin appearance, and expression. We also compare with competing techniques in each analysis.

\subsubsection{Light Distance}
\begin{figure}[t]
\begin{center}
   \includegraphics[width=1\linewidth]{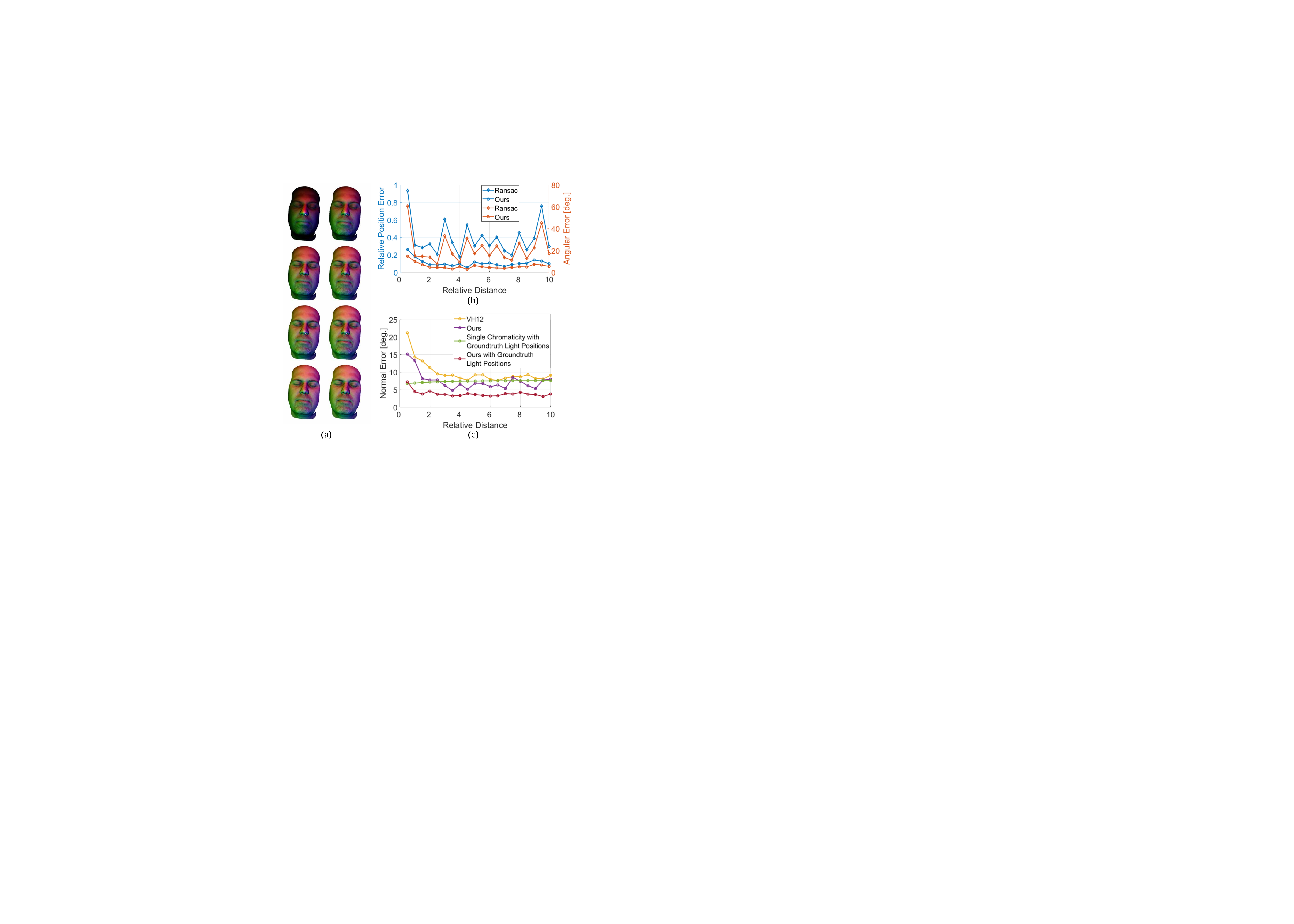}
\end{center}
   \caption{Effect of changing light source distances. (a) Rendered images under the first 8 distances (distance increases from left to right and from top to bottom). Comparisons on (b) self-calibration and (c) reconstructed normal errors at different light source distances, including against VH12~\cite{vogiatzis2012self}.}
\label{fig:result_dist_err_stats}
\end{figure}

In this experiment, we vary the distance between the light sources and face mesh while fixing the elevation angle at $65\degree$. The distance is specified in terms of vertical span of the face; it ranges from 0.5 to 10 with increments of 0.5. The rendered images for the first 8 distances are shown in Fig.~\ref{fig:result_dist_err_stats}a.

Fig.~\ref{fig:result_dist_err_stats}b compares the calibration errors for vanilla RANSAC and our method.
We first transform the calibration results to the same coordinate system as the groundtruth light positions before computing errors.
We compute the relative position error as Euclidean position error normalized by light source distance. The angular error is computed with regard to the face center. We can see that vanilla RANSAC is less accurate with large fluctuations in error over distance. By comparison, our calibration results are more accurate and robust to changing light source distance, with the relative position error around 0.1 and the angular error around $5\degree$ for most distances. 

\begin{figure}[t]
\begin{center}
   \includegraphics[width=0.9\linewidth]{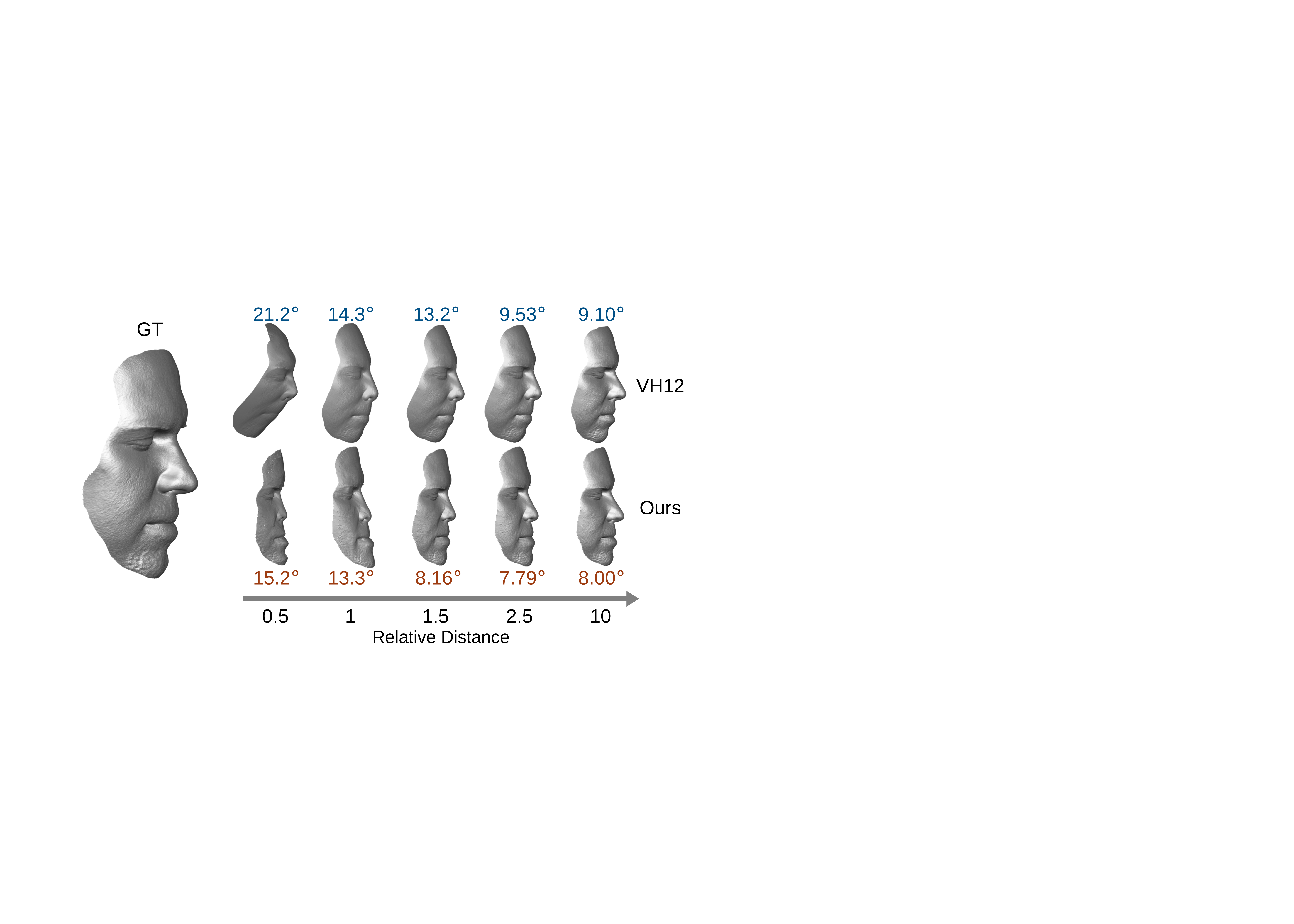}
\end{center}
   \caption{Comparison of reconstructed geometry with VH12~\cite{vogiatzis2012self} at different light source distances. The colored numbers are the mean normal errors.}
\label{fig:result_dist_recon}
\end{figure}

We also compare the reconstruction accuracy of our method using our estimated light positions with VH12 \cite{vogiatzis2012self} in Fig.~\ref{fig:result_dist_err_stats}c. 
% \cite{vogiatzis2012self} assumes directional lighting with single albedo chromaticity, and uses the same proxy face for self-calibration. 
We can see that our method consistently performs better, even at distance 10 (where lighting is almost directional). There is considerable shape deformation for \cite{vogiatzis2012self} across the different distances as shown in Fig.~\ref{fig:result_dist_recon}, while our method produces reasonable shapes starting from distance 1.5. At very close distances such as 0.5 and 1, both methods do not perform well due to significant self-shadowing. 

Fig.~\ref{fig:result_dist_err_stats}c also shows comparisons with using ground truth light positions and mean albedo chromaticity (which are the conditions that should result in the best accuracy under the single chromaticity assumption). In this case, our method using ground truth light positions out-performs the others by a significant margin under almost all distances; this shows the importance of spatially-varying albedo chromaticity. The degraded accuracy at distance 0.5 is due to significant self-shadowing.

\subsubsection{Light Elevation Angle}

\begin{figure}[t]
\begin{center}
   \includegraphics[width=0.6\linewidth]{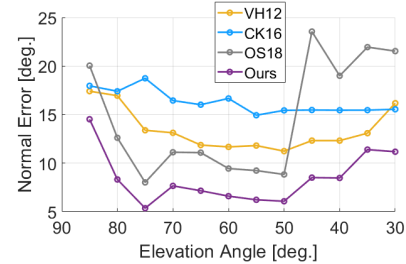}
\end{center}
   \caption{Mean normal errors (in degrees) under different light elevation angles, compared with VH12~\cite{vogiatzis2012self}, CK16~\cite{chakrabarti2016single}, and OS18~\cite{ozawa2018single}.}
\label{fig:result_angle_err_stats}
\end{figure}

The elevation angle of light sources have a direct impact on light source baseline. A large elevation angle results in a small light source baseline, which enables the equipment to be more portable. However, the angular difference between light sources decreases as the elevation angle increases, which in turn makes reconstruction less robust. 
In the extreme case where the elevation angle is $90\degree$, the three light sources degenerate into a single light source, with their spectra combined. On the other hand, small elevation angles results in more self-shadowing, which also negatively impacts reconstruction.

We fix the distance at 2.0 and vary the elevation angle from $85 \degree$ to $30 \degree$ with a decrement of $5 \degree$. At the elevation angle of $30 \degree$, about $30\%$ of facial pixels are in shadow for green and blue lights. Fig.~\ref{fig:result_angle_err_stats} shows the mean normal error of our method against~\cite{vogiatzis2012self,chakrabarti2016single,ozawa2018single}. It can seen that our method consistently performs the best under all elevation angles. In addition, all methods display a trend to perform worse at two ends of elevation angles, although \cite{chakrabarti2016single} is less affected by severe self-shadowing at small elevation angle. While \cite{ozawa2018single} produces smaller errors than \cite{vogiatzis2012self,chakrabarti2016single} under medium elevation angles, its accuracy drastically degrades for small elevation angles due to self-shadowing. This highlights the importance of our proposed albedo similarity and proxy prior in correcting errors led by albedo consensus.

\begin{figure*}
\begin{center}
   \includegraphics[width=0.9\linewidth]{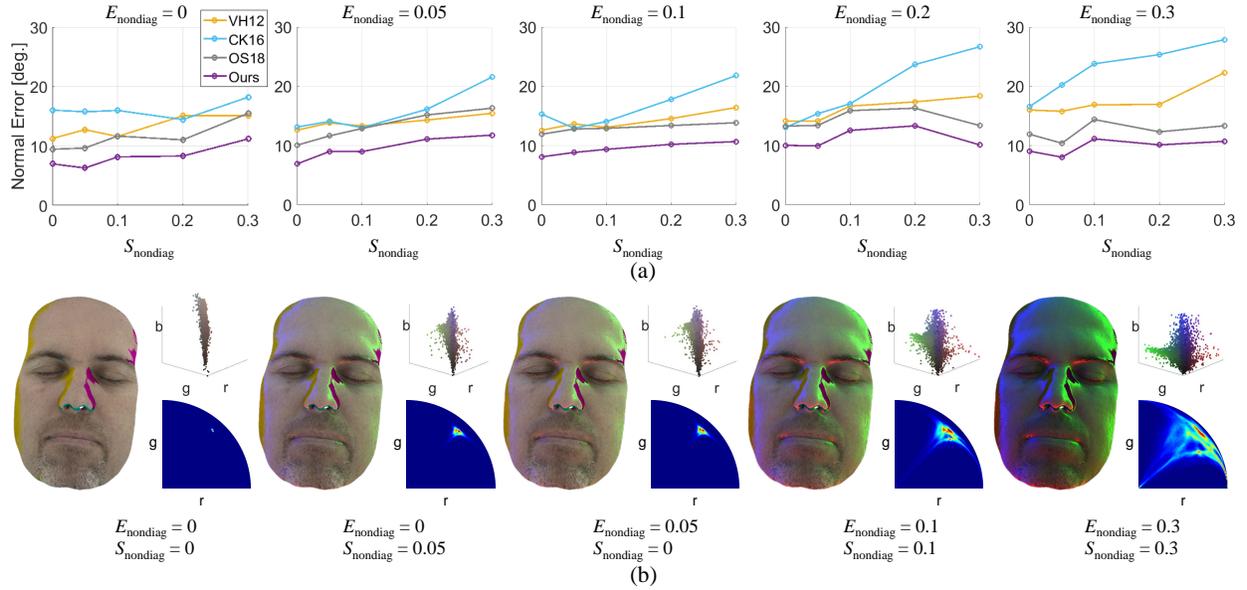}
\end{center}
   \caption{Experimental results for different crosstalk parameters $E_{\rm nondiag}, S_{\rm nondiag}$.
   (a) Mean normal errors (in degrees), compared with VH12~\cite{vogiatzis2012self}, CK16~\cite{chakrabarti2016single}, and OS18~\cite{ozawa2018single}. (b) Apparent albedo maps and albedo/albedo chromaticity distributions for five representative examples. 
   The albedo values are rescaled so that most pixels are within $[0, 1]$. (Albedo values at pixels near shadows can be very large; they are not used for rescaling.)
   }
\label{fig:result_crosstalk_err_stats}
\end{figure*}

\begin{figure}[t]
\begin{center}
   \includegraphics[width=0.9\linewidth]{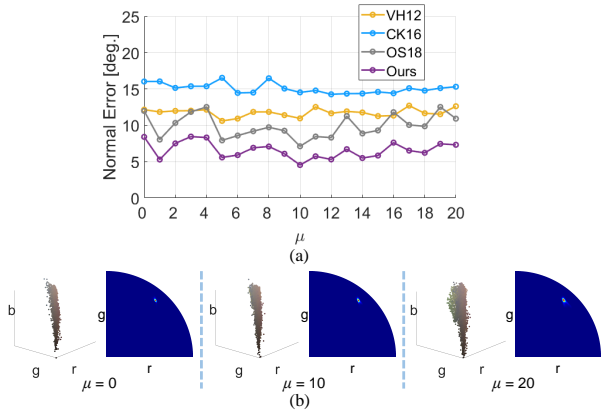}
\end{center}
   \caption{Experimental results for different anisotropy parameter $\mu$. (a) Mean normal errors (in degrees), compared with VH12~\cite{vogiatzis2012self}, CK16~\cite{chakrabarti2016single}, and OS18~\cite{ozawa2018single}. (b) Albedo distribution (left) and albedo chromaticity distribution (right) for $\mu = 0, 10, 20$.}
\label{fig:result_anisotropy_err_stats}
\end{figure}

\begin{figure*}
\begin{center}
   \includegraphics[width=1\linewidth]{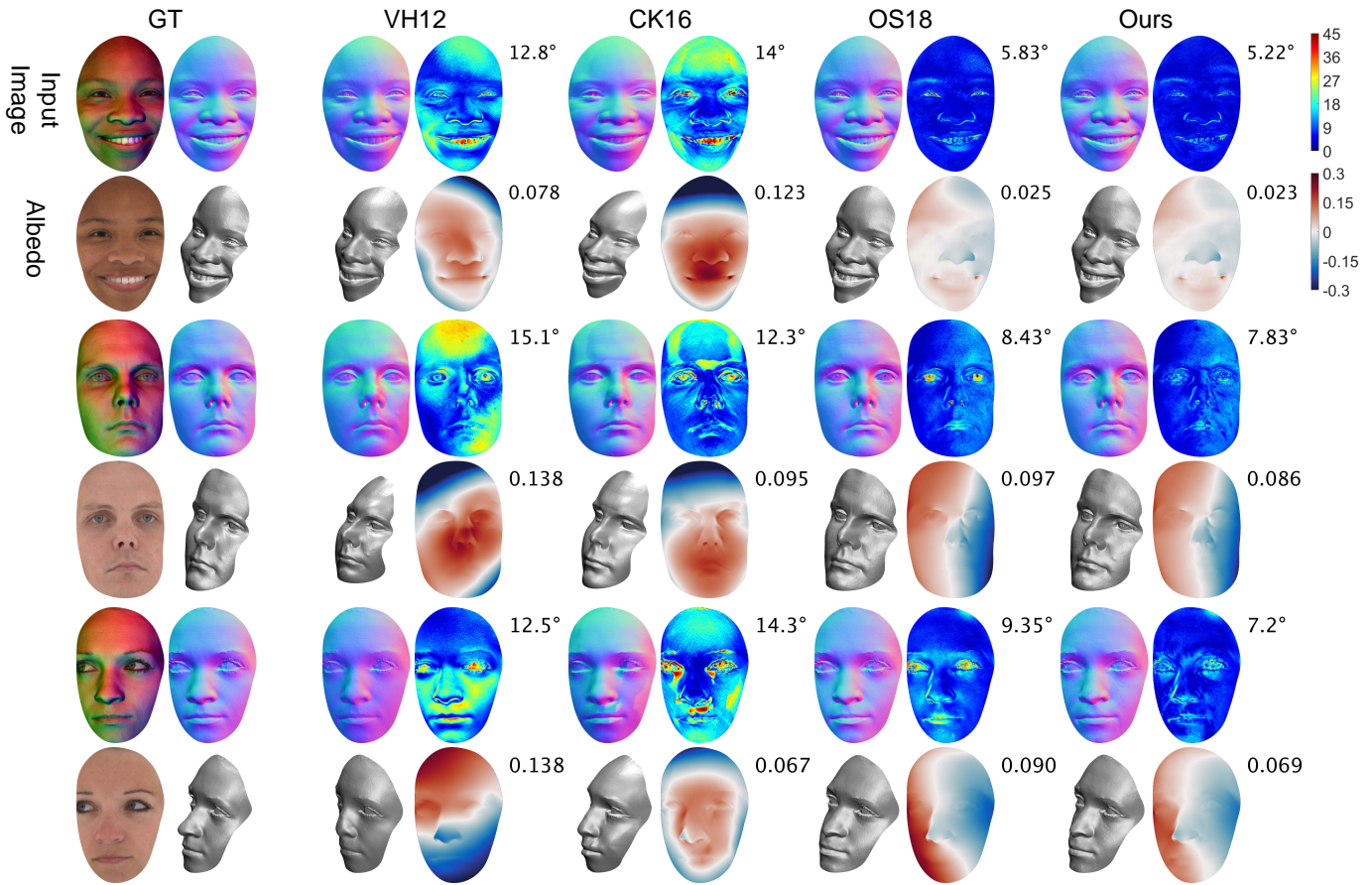}
\end{center}
   \caption{Comparison with competing techniques (VH12~\cite{vogiatzis2012self}, CK16 \cite{chakrabarti2016single}, OS18 \cite{ozawa2018single}) using data from the ICT-3DRFE dataset. GT is ground truth. The mean normal and geometry errors are listed in the odd and even rows, respectively. More results can be found in the supplementary file.}
\label{fig:result_3drfe_recon}
\end{figure*}

\subsubsection{Light Anisotropy}
Unlike the ideal point light model, real LEDs exhibit anistropic intensity patterns. To analyze its effect on our method, we further render images using an anisotropic point light model \cite{queau2018led}:
\begin{equation}\label{anisotropic_light}
\mathbf{c}_i(x,y) = \boldsymbol{\rho}_i(x,y)(\frac{\mathbf{n}_s^i\cdot\mathbf{L}_i(x,y)}{\left\lVert\mathbf{L}_i(x,y)\right\rVert_2})^{\mu^i}\mathbf{L}_i(x,y)\mathbf{n}(x,y),
\end{equation}
where $\mathbf{n}_s^i$, $\mu^i$ are the (unit-length) principal direction and anisotropy parameter of the $i$th light source. The anisotropy parameter equals $0$ for ideal point light source while a larger value indicates stronger radial attenuation around the principal direction. We render the images at distance 2.0 and elevation angle $65 \degree$, with anisotropy parameter ranging from 0 to 20. With $\mu = 20$, the half-intensity angle is only about $15 \degree$, revealing very strong radial attenuation. 

As shown in Fig.~\ref{fig:result_anisotropy_err_stats}(a), light anisotropy has no noticeable adverse effect on both our method and comparison methods. We further compute apparent albedo under different anisotropy parameters by using Eq.~\ref{eq:near_color_ps_matrix_simple} along with ground truth light position, normal, and per-pixel 3D position. (Note that the apparent albedo is not the true albedo, since it incorporates any outlier effect such that the rendering equation adheres to Eq. \ref{eq:near_color_ps_matrix_simple}.) Here, light anisotropy is entirely incorporated as part of albedo. Fig.~\ref{fig:result_anisotropy_err_stats}(b) shows the albedo/albedo chromaticity distributions for three cases, where there is only minor difference. This illustrates why light anisotropy has little influence on the accuracy of all methods and validates our use of ideal point light model.

\subsubsection{Crosstalk}
Similar to light anisotropy, the influence of crosstalk can also be interpreted as modification on albedo/albedo chromaticity distribution. To simulate crosstalk, due to a lack of hyperspectral reflectance data, we only consider the wavelengths corresponding to red, green, and blue when evaluating Eq.~\ref{eq:albedo_matrix}, which can be rewritten as:
\begin{equation}\label{eq:albedo_matrix_simple}
\mathbf{A}(x,y)=\mathbf{S}\text{diag}(\boldsymbol{r}(x,y))\mathbf{E},
\end{equation}
where $\mathbf{S}, \mathbf{E}\in\mathbb{R}^{3\times3}$, $\boldsymbol{r}(x,y)\in\mathbb{R}^{3\times1}$ and $\mathbf{S}_{i,k}=\mathcal{S}_i(\lambda_k)$, $\mathbf{E}_{k,j}=\mathcal{E}_j(\lambda_k)$, $\boldsymbol{r}_k(x,y)=\mathcal{R}(x,y,\lambda_k)$. Crosstalk exists when any non-diagonal element of $\mathbf{S}$, $\mathbf{E}$ is non-zero. We set the diagonal elements to $1$ and gradually increase their non-diagonal elements $S_{\rm nondiag}, E_{\rm nondiag}$ (non-diagonal elements are set as the same) to simulate increasing crosstalk.
%\singbing{All the non-diagonal elements are the same? Or did you randomize the numbers such that the parameters reflect the standard deviation? The latter would probably make more sense.}

Fig.~\ref{fig:result_crosstalk_err_stats}(a) shows the normal errors under different combinations of $\mathbf{S}$ and $\mathbf{E}$, where generally more crosstalk leads to worse accuracy for all methods. Fig.~\ref{fig:result_crosstalk_err_stats}(b) shows the apparent albedo maps along with albedo/albedo chromaticity distributions for 5 cases. We can see that with more crosstalk, there is stronger spatial albedo variation, which violates the piecewise constancy assumption of \cite{chakrabarti2016single}. Although \cite{vogiatzis2012self} does not have a no-crosstalk requirement, it also significantly suffers from the spreading-out of albedo chromaticity distribution due to crosstalk. Our method is more robust to this phenomenon because of the incorporation of albedo similarity and proxy prior.

\subsubsection{Evaluation on ICT-3DRFE}

% xxx: results using gt light position for ozawa and our method

\begin{figure}[t]
\begin{center}
   \includegraphics[width=1\linewidth]{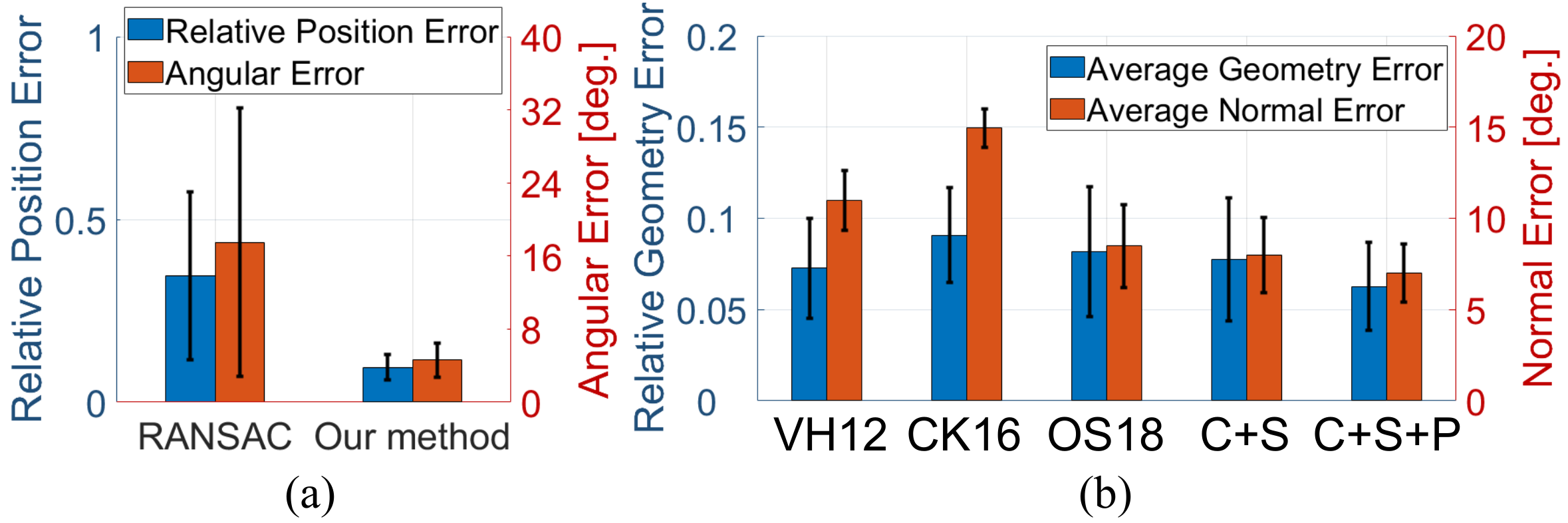}
\end{center}
   \caption{Error statistics on ICT-3DRFE dataset. (a) Self-calibration errors. (b) Reconstruction errors of VH12~\cite{vogiatzis2012self}, CK16~\cite{chakrabarti2016single}, OS18~\cite{ozawa2018single}, our ``Consensus + Similarity" and our ``Consensus + Similarity + Proxy".}
\label{fig:result_3drfe_err_stats}
\end{figure}

\begin{figure*}
\begin{center}
   \includegraphics[width=0.9\linewidth]{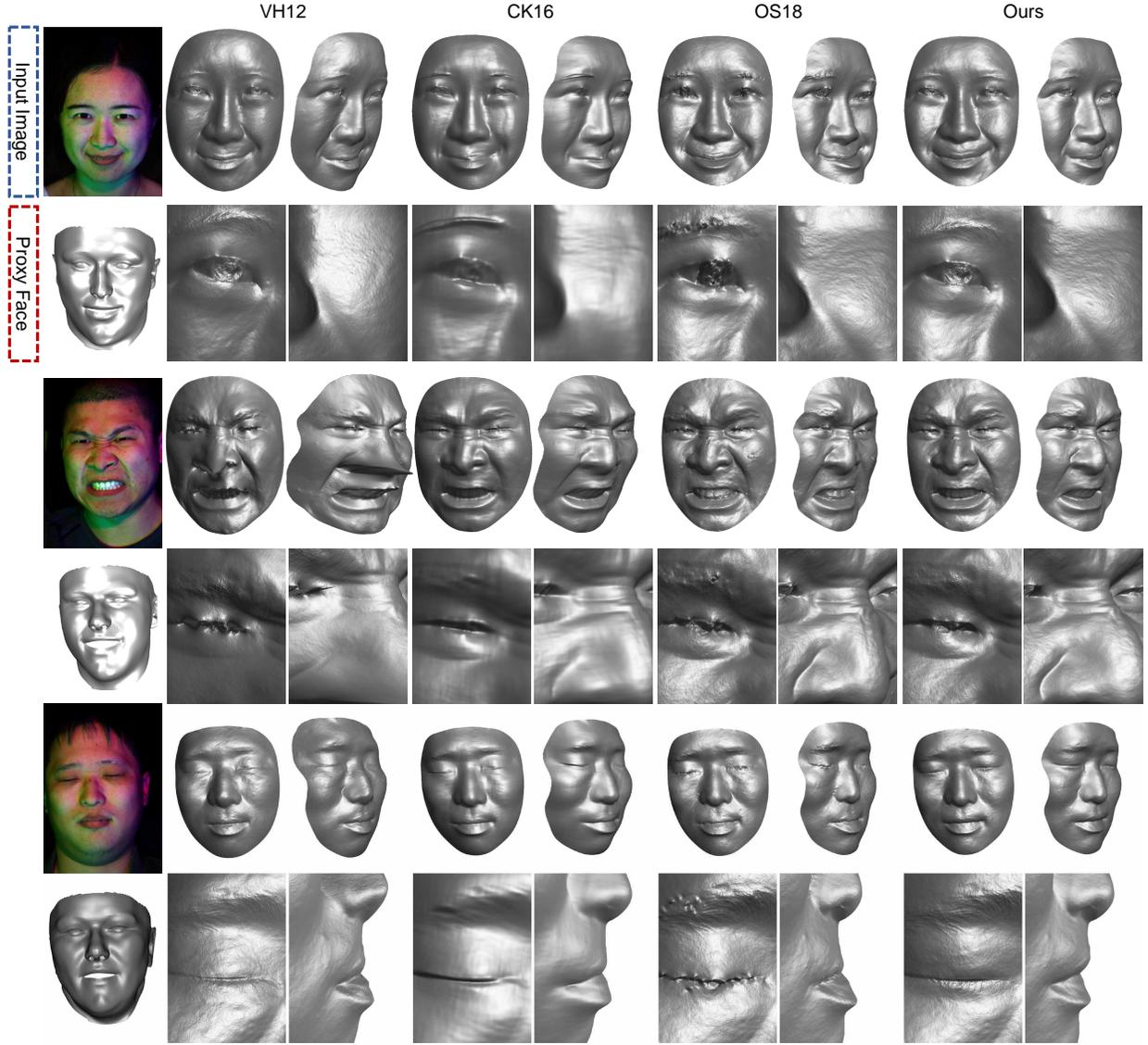}
\end{center}
   \caption{Reconstruction results of VH12~\cite{vogiatzis2012self}, CK16~\cite{chakrabarti2016single}, OS18~\cite{ozawa2018single} and our method on real data. More examples are in the supplementary file.}
\label{fig:result_real_recon}
\end{figure*}

We further evaluate our method using the ICT-3DRFE dataset~\cite{stratou2011effect}, which contains highly-detailed albedo and geometry for 23 subjects (22 with 15 expressions each, and one with 11 expressions, with a total of 341 face inputs). The dataset has vastly different skin reflectance as well as face geometry. We rendered images at light source distance $2.0$ and elevation angle $65 \degree$, with no anisotropy or crosstalk. We also added Gaussian noise ($\sigma_{\rm noise} = 2/255$) to simulate real images. 
%Results are shown in Fig.~\ref{fig:result_3drfe_err_stats}.

As shown in Fig.~\ref{fig:result_3drfe_err_stats}a, our self-calibration method significantly improves over vanilla RANSAC. In Fig.~\ref{fig:result_3drfe_err_stats}b, we compare the accuracy of our face reconstruction method with those of \cite{vogiatzis2012self,chakrabarti2016single,ozawa2018single}. For our method, we show results of two variants (``Consensus + Similarity" and ``Consensus + Similarity + Proxy") to analyze the influence of each term. 
%For \cite{ozawa2018single} and the two variations of our method, we use the light positions estimated in our self-calibration. 
We compute relative geometry error as depth error of integrated geometry normalized by depth range of ground truth geometry. Methods using near point light model outperform those using directional light model in terms of normal error. Each proposed term improves over using consensus only. 

Although \cite{chakrabarti2016single} handles multi-chromaticity, it performs worse than \cite{vogiatzis2012self}. It is likely that its polynomial model for depth is not suitable for complex geometry. While \cite{vogiatzis2012self} has lower geometry error than using consensus only \cite{ozawa2018single}, our full method improves over this metric and yields the best accuracy. Fig.~\ref{fig:result_3drfe_recon} shows 3 %(xxx: modify if figure changes)
comparisons. Our method works reasonably well in the lip and eyebrow regions, even though they contain non-dominant albedos (which tend to cause incorrect consensus). Shadows, as with light anisotropy and crosstalk, can also be explained by apparent albedo; they result in additional albedo variation (see the leftmost albedo map in Fig.~\ref{fig:result_crosstalk_err_stats}(b)). Since our formulation does not enforce spatial constancy, it can better handle such variation compared with \cite{vogiatzis2012self,chakrabarti2016single}. Still, our reconstructions contain errors at shadowed regions near the nose due to inaccurate proxy mesh around the nose. Please see the supplementary material for detailed error statistics and more results.

\subsection{Experiments Using Real Data}
\begin{figure}[t]
\begin{center}
   \includegraphics[width=1\linewidth]{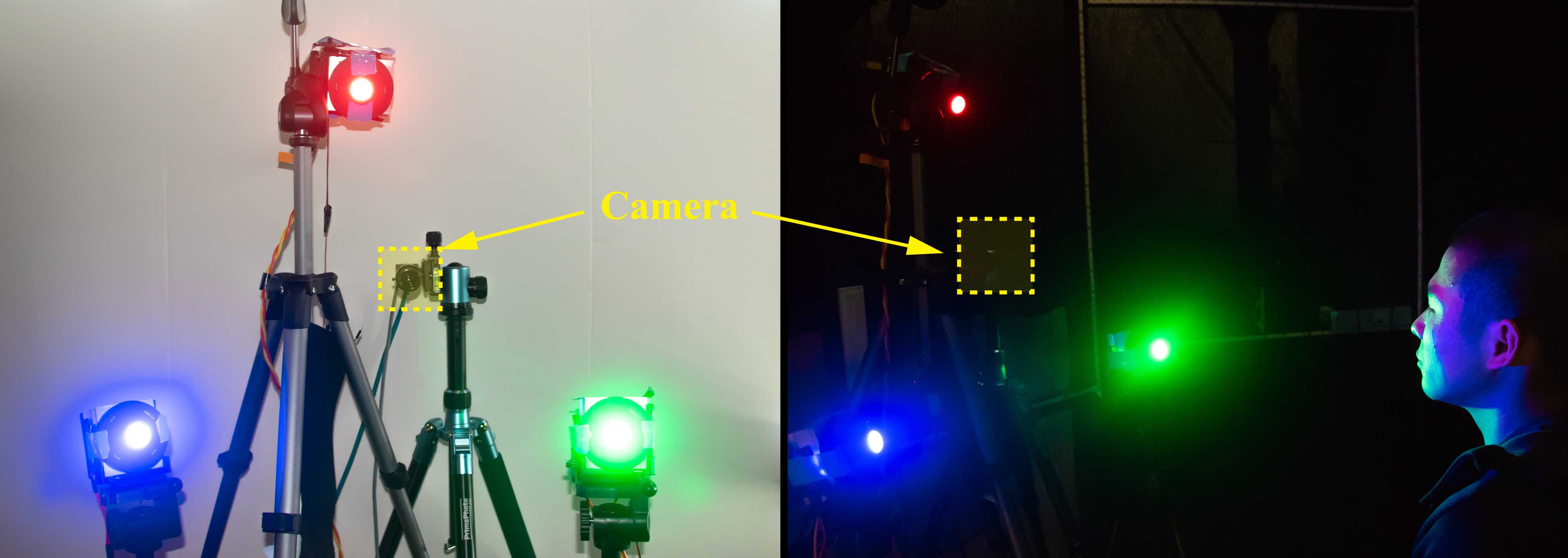}
\end{center}
   \caption{Hardware setup of our capture system.}
\label{fig:hardware_setup}
\end{figure}

To collect real data, we built a color photometric capture system as shown in Fig.~\ref{fig:hardware_setup}. It consists of 3 LED (red, green, blue) near point lights and a PointGrey Flea3 FL3-U3-88S2C color camera ($4096 \times 2160$). The distance between the light sources and subject is roughly 70cm. We mounted orthogonal linear polarizers in front of the light sources and camera to reduce specular reflection. 

To reduce crosstalk, we compute a de-crosstalk matrix from three images of a white paper, namely, one image for each of the three lights. (This step is done only once.) Specifically, the de-crosstalk matrix is computed as
\begin{gather}\label{eq:decrosstalk}
 \mathbf{M} =
  \begin{bmatrix}
   1 & \text{med}(I_r^g \oslash I_g^g) & \text{med}(I_r^b \oslash I_b^b) \\
   \text{med}(I_g^r \oslash I_r^r) & 1 & \text{med}(I_g^b \oslash I_b^b) \\
   \text{med}(I_b^r \oslash I_r^r) & \text{med}(I_b^g \oslash I_g^g) & 1
   \end{bmatrix}^{-1},
\end{gather}
where $I_r^g$ is the red channel of the image under green light, $\oslash$ is Hadamard division operator and $\text{med}(\cdot)$ yields the median value. This matrix is left multiplied with the RGB value of each pixel.

The final mesh consists of about 3,000,000 vertices. The whole process takes about 12 minutes on a 6-core 3.7GHz CPU with 64GB memory, whereas self-calibration takes 8 minutes and face reconstruction takes 4 minutes. Same with synthetic experiments, proxy faces are made available to \cite{vogiatzis2012self} for self-calibration while \cite{chakrabarti2016single,ozawa2018single} are provided with calibration information.

% xxx: calibration errors of VH12 and ours
\begin{figure}[t]
\begin{center}
   \includegraphics[width=1\linewidth]{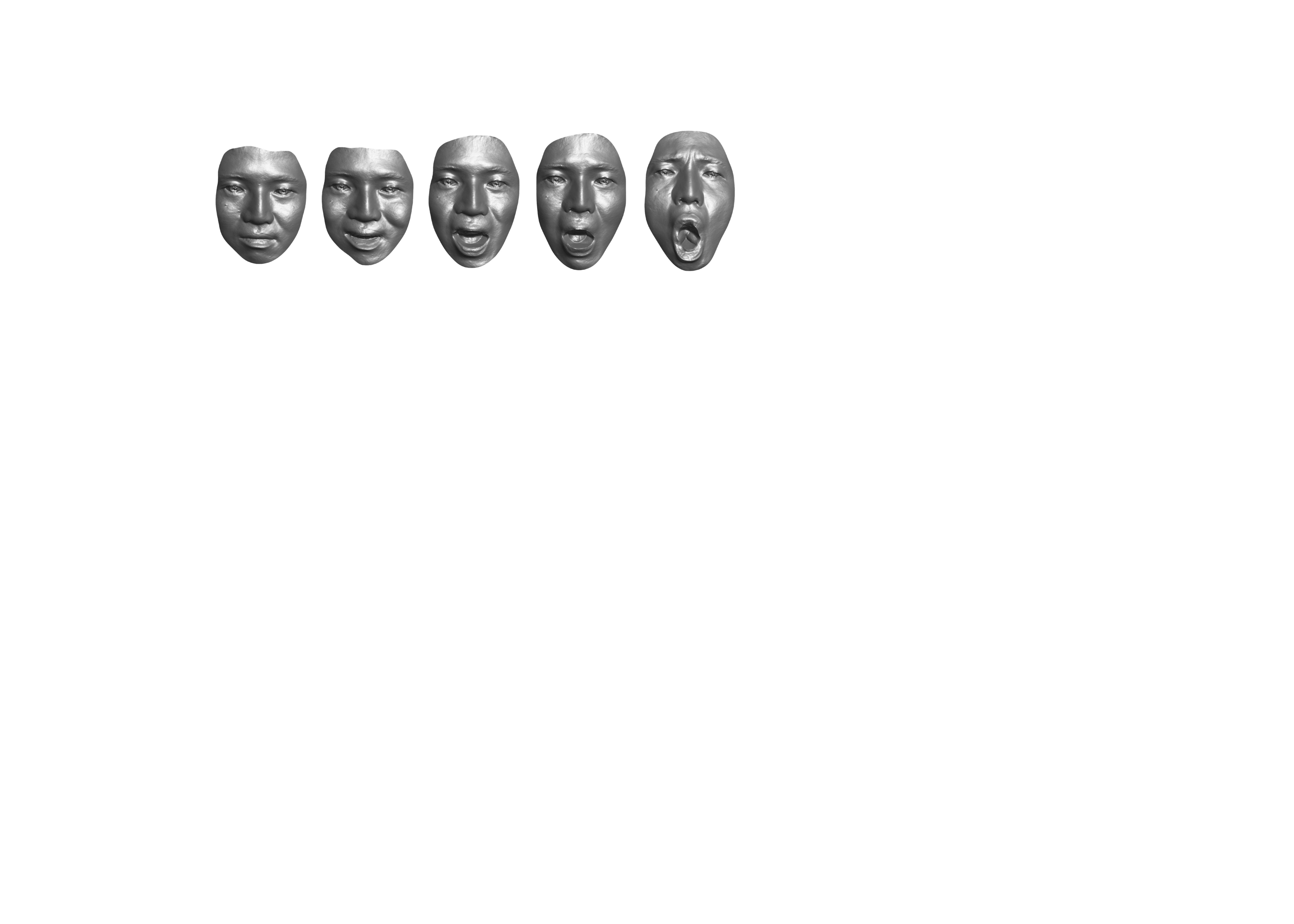}
\end{center}
   \caption{Reconstruction results for a video clip of a face with changing expressions. Each frame is processed independently.}
\label{fig:result_real_dynamic_recon}
\end{figure}

We captured faces of different people and expressions; Fig.~\ref{fig:result_real_recon} shows results for 3 %(xxx: modify if figure changes)
examples (including different gender and expressions). 
Results from competing techniques (VH12~\cite{vogiatzis2012self}, CK16~\cite{chakrabarti2016single}, and OS18~\cite{ozawa2018single}) feature local and global geometric distortion as well as over-smoothing. These results also have issues at the lips, and this is because the albedo at the lips differ from those at the rest of the face. Notice that our method works well for exaggerated expressions (such as the second example) even though the proxy face does not accurately depict the expression.
% Common to all methods is the distortions due to shadows, especially around the nose. Still, our method provides 
Please refer to the supplementary material for more results. 

We have also captured a video clip of a face with changing expressions and reconstructed each frame independently. Fig.~\ref{fig:result_real_dynamic_recon} shows results for 5 representative frames. The mouth interior was not reconstructed well due to significant self-shadowing.

\begin{figure}[t]
\begin{center}
   \includegraphics[width=0.825\linewidth]{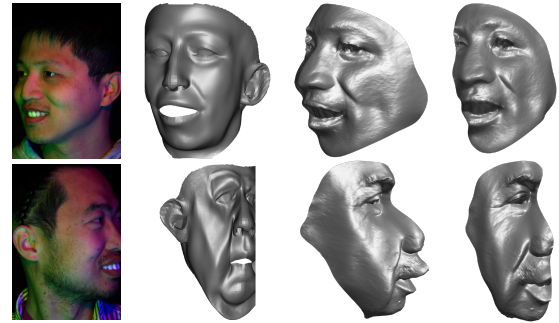}
\end{center}
   \caption{Failure cases for our method. The first two columns are input image and proxy face, while the last two columns are two views of our reconstruction.}
\label{fig:result_real_failure}
\end{figure}

Fig.~\ref{fig:result_real_failure} shows two failure cases for our method, which contain extreme poses. The reason is that the proxy face generated by 3DMM fitting is significantly less accurate under such poses. This affects our algorithm in two ways: (1) self-calibration of light sources is less robust due to significant pose misalignment of proxy face, and (2) highly incorrect proxy normals adversely affect face reconstruction due to incorrect proxy term.

%\singbing{Show a picture of the setup. Mention frame rate, image resolution.}

%\singbing{Don't forget to mention time performance, with timing for different steps. There should also be mention of image resolution, mesh size, etc.}

%\singbing{What are the failure modes? Show examples.}

%\singbing{Discuss how 3D facial dynamics capture can be improved with temporal coherency.}

\section{Conclusion}

We have presented a novel color photometric stereo (CPS) method with only 3 uncalibrated near point lights. Our method is capable of reconstructing high-quality face geometry from {\em a single image}. Self-calibration of the near point lights relies on the geometric prior from the 3DMM proxy face. We apply RANSAC, followed by hypothesis merging to robustly estimate light positions. We also propose a per-pixel formulation for reconstruction that incorporates albedo consensus, albedo similarity, and proxy prior to handle the ill-posedness of CPS. Synthetic and real experiments show that our method outperforms previous CPS methods that similarly use a single image as input.

In our work, we did not exploit the albedo prior of human faces; this prior may further improve the accuracy of self-calibration and face reconstruction. While not trivial, it would also be interesting to explicitly handle self-shadows. Another possible future work would be extending our method to general objects by learning from depth sensor observations.

% Any acknowledgments to only be included in camera ready
\ifpeerreview \else
% \section*{Acknowledgments}
\fi

\bibliographystyle{IEEEtran}
\bibliography{references}

\ifpeerreview \else
%%%% For the camera ready version, please fill out this
%%%% biography. Your camera ready should be within a 12 page limit
%%%% including acknowledgments, references and biography.

% \includegraphics command within an optional argument. (You could
% create your own custom macro containing the \includegraphics command
% to make things simpler here.)
% \begin{IEEEbiography}[{\includegraphics[width=1in,height=1.25in,clip,keepaspectratio]{mshell}}]{Michael Shell}

% \end{IEEEbiography}

%\newpage


%\vfill

\fi

\end{document}